\documentclass{article}

\usepackage{microtype}
\usepackage{graphicx}
\usepackage{subfigure}
\usepackage{booktabs} 

\usepackage{enumitem}
\usepackage{dblfloatfix}
\usepackage{multirow}
\usepackage{soul}

\usepackage{hyperref}

\usepackage[accepted]{icml2019}

\icmltitlerunning{Paradox in Deep Neural Networks: Similar yet Different while Different yet Similar}

\begin{document}

\def\eg{\emph{e.g.}}
\def\ie{\emph{i.e.}}
\def\Eg{\emph{E.g.}}
\def\etal{\emph{et al.}}
\def\etc{\emph{etc.}}
\def\aka{\emph{a.k.a. }}

\twocolumn[
\icmltitle{Paradox in Deep Neural Networks: \\
           Similar yet Different while Different yet Similar}




\begin{icmlauthorlist}
\icmlauthor{Arash Akbarinia}{gie}
\icmlauthor{Karl R. Gegenfurtner}{gie}
\end{icmlauthorlist}

\icmlaffiliation{gie}{Abteilung Allgemeine Psychologie, Justus-Liebig-Universit\"at, Giessen, Germany}

\icmlcorrespondingauthor{Arash Akbarinia}{arash.akbarinia@psychol.uni-giessen.de}

\icmlkeywords{Machine Learning, ICML}

]



\printAffiliationsAndNotice{\icmlEqualContribution} 

\begin{abstract}
Machine learning is advancing towards a data-science approach, implying a necessity to a line of investigation to divulge the knowledge learnt by deep neuronal networks. Limiting the comparison among networks merely to a predefined intelligent ability---according to ground truth---does not suffice, it should be associated with innate similarity of these artificial entities. Here, we analysed multiple instances of an identical architecture trained to classify objects in static images (\textit{CIFAR} and \textit{ImageNet} data sets). We evaluated the performance of the networks under various distortions and compared it to the intrinsic similarity between their constituent kernels. While we expected a close correspondence between these two measures, we observed a puzzling phenomenon. Pairs of networks whose kernels' weights are over 99.9\% correlated can exhibit significantly different performances, yet other pairs with no correlation can reach quite compatible levels of performance. We show implications of this for transfer learning, and argue its importance in our general understanding of what intelligence is---whether natural or artificial.
\end{abstract}

\section{Kernel-science}
\label{intro}

Artificial intelligence is evolving at a great rate, surpassing our very own in various fields, from the famous \textit{AlphaZero}, a generic model conquering the world of intellectual board games, such as, go, chess and shogi, \cite{silver2018general}, to a \textit{3D-CNN} that swiftly reports abnormalities in medical images \cite{titano2018automated}. Despite these impressive achievements, our understanding of the underlying mechanisms in deep networks---the fundamental technology behind the scenes---lags far behind.

Deciphering the basis of intelligence requires adequate explanations to, at least, two principal questions:
\begin{enumerate}[label=\roman*]
    \item \textbf{What} operations lead to intelligent behaviour?
    \item \textbf{How} can an intelligent behaviour be accomplished?
\end{enumerate}
It is worth acknowledging that a resolution to the latter does not necessarily shed light on the former. For instance, conventional wisdom \cite{lecun2015deep} states that given a large enough amount of labelled data, deep neuronal networks (DNN) can accurately learn a task (the \textbf{how} question). At the same time, we do not possess a clear understanding of why DNNs work so excellently \cite{zeiler2014visualizing} (the \textbf{what} question). This is one of the biggest challenges the artificial intelligence community is facing at present.

A more thorough comprehension of artificial networks is of great importance from several perspectives:
\begin{enumerate}[label=\roman*]
    \item \textit{Performance improvement}. Many elements and their constituent hyper-parameters, such as network architecture, optimisation algorithm, data augmentation, weights initialisation, \etc, greatly influence the outcome of a training procedure. Currently, the performance boost is achieved through cumbersome parameter tuning \cite{glorot2010understanding}. A better understanding of what operations are executed to represent a feature inside a network \cite{bengio2013representation} would facilitate this procedure.
    \item \textit{Critical applications}. DNNs are becoming a ubiquitous tool in various systems. However, they suffer from the butterfly effect: tiny variations in input cause dramatic changes in output, \eg, \cite{stallkamp2012man, szegedy2013intriguing, moosavi2016deepfool, tian2018deeptest}, this, in turn, implies vulnerability to malicious attacks \cite{papernot2017practical}. Therefore, trusting DNNs with more critical applications, such as, monitoring nuclear plants or automatically diagnosing patients, is subject to a more profound insight into their operations.
    \item \textit{Human perception}. Recent developments in machine learning have created an amazing opportunity to gain more awareness about the neuronal network inside our brain \cite{marblestone2016toward}. DNNs are actively studied as models of our perception, \eg, vision \cite{eckstein2017humans, flachot2018processing}, hearing \cite{kell2018task}, touch \cite{zhao2018neural}, to name a few. Much of these works happen at a coarse level---a population of cells. Deciphering kernels' operations could elucidate finer details of our own brain---individual neurons.
\end{enumerate}

\subsection{Related works}

This article attempts to discuss the \textit{what} question in the context of visual information processing. Previous works on this topic could be broadly summarised into three groups:
\begin{enumerate}[label=\roman*]
    \item \textit{Visualisation}. A large body of literature is devoted to visualising internal units of DNNs, \eg, \cite{simonyan2013deep,zeiler2014visualizing,mahendran2015understanding}. Despite their genuine usefulness to give an idea of what kernels respond to, they are of a qualitative nature and should be complemented with quantitative techniques.
    \item \textit{Transfer learning}. Another set of papers investigates transferability of knowledge across networks, data sets and tasks, \eg, \cite{agrawal2014analyzing, yosinski2014transferable}. Although, they empirically demonstrate the crucial hierarchical characteristics of layers, \ie, the transition from generic to specific in deeper layers, thus far, no account of feature invariance has been provided.
    \item \textit{Cognitive}. Further works attempt to interpret the intrinsic behaviour of kernels by analysing their activation patterns, \eg, \cite{Bau_2017_CVPR,akbarinia2018contrast,zhang2018interpretable}. These techniques successfully exhibit the existence of selectivity among kernels, similar to biological neurons \cite{quiroga2005invariant}. However, the causality of these kernels for a specific function remains to be demonstrated.
\end{enumerate}

\subsection{Research hypothesis}

Our approach consists of generating two groups of networks. First, we trained multiple instances of an identical architecture under distinct conditions, ultimately aiming to classify objects in natural images. Second, we fine-tuned each model with various types of distortions and transformations, typically experienced by an intelligent agent. We computed the classification accuracy of all networks for eight types of image manipulation, such as contrast reduction or adding noise. This measure is considered the generalised cognitive capacity of a network. Next, we calculated a statistical metric of intrinsic similarity between the kernels' weights. Our \textit{research hypothesis} states that those instances whose overall accuracy is alike would tend to be of a more similar nature and vice versa. Networks whose weights are nearly identical would show a comparable level of performance.

It could be objected that in the hyper-dimensional space of neural networks, such direct correspondences between kernels are meaningless. Nevertheless, a large portion of our knowledge about biological intelligence, in neuroscience, has arguably been acquired through such neuron to neuron comparisons. For instance, a recent study of single neuron comparison has provided important insights about the differences between humans and monkeys brains in terms of the trade-off between ``robustness'' and ``efficiency'' \cite{pryluk2019tradeoff}. Furthermore, given kernels' activation maps are often compared across networks to an interpretable concept, \eg, \cite{Bau_2017_CVPR,zhang2018visual}, therefore, it is expected that the underlying parameters of those kernels to be comparable as well.

\subsection{Contribution}

In total we trained 329 instances of an identical architecture---\textit{ResNet20} \cite{he2016deep}---to classify objects in the \textit{CIFAR} data set \cite{krizhevsky2009learning}. We analysed every single pair of these networks, \ie, 53,956 comparisons ($\frac{329 \times 328}{2}$). We also trained 20 instances of \textit{ResNet50} on \textit{ImageNet} data set \cite{krizhevsky2012imagenet}, resulting in comparing 190 pairs of networks.

The results of our experiments refute the research hypothesis. On the one hand, we encountered numerous pairs of networks with a great resemblance in their performance, although their constituent kernels show no similarity whatsoever. On the other hand, the behaviour of many other pairs of networks is very different, despite the fact that their kernels are almost identical.

We further investigated whether a direct copy of weights from one network to another could convey the associated cognitive capacities as well. The results of our experiment suggest that such a na\"ive transfer learning is effective for those instances with high intrinsic similarity. 
This, in turn, strengthens the second part of our hypothesis: as networks' kernels become more similar, so does their behaviours.

\section{Methods}

\subsection{Data set}

We conducted our experiments on two benchmark data sets of object classification in static images:
\begin{itemize}
    \item \textit{CIFAR} \cite{krizhevsky2009learning} consists of 60 thousands colour images of size 32 $\times$ 32, belonging to 10 classes of objects. The training and validation sets are divided by a ratio of 5 to 1, respectively. 
    \item \textit{ImageNet} \cite{krizhevsky2012imagenet} is a collection of 1,000 object categories. The training and validation set contain 1.3 million and 50 thousands images, respectively (\ie, 1300 and 50 per category).
\end{itemize}

\subsection{Network visual intelligence}
\label{sec:congnition}
We defined the cognitive visual capacity of a network as its overall accuracy under eight types of image manipulation that are common and therefore of great importance for machine intelligence:
\begin{enumerate}
    \item \textit{Contrast reduction}. We modulated contrast through this equation: $I_c^{i}(x,y) = \frac{c}{100} \times I^{i}(x,y) + \frac{1 - \frac{c}{100}}{2}$, where $I$ is the input image, $\left\{x,y\right\}$ are pixel coordinates, $i$ denotes colour channel, and $c$ is the contrast level.
    \item \textit{Illuminant variation}. We simulated this by altering the ratio of colour channels: $I_l^{i}(x,y) = I^{i}(x,y) \times l^{i}$, where $l$ represents the luminance of a scene. For instance, $l=\left[0,1,0\right]$ results in a completely blue image.
    \item \textit{Image blurring}. We blurred an image through its convolution with a Gaussian function: $I_\sigma^{i}(x,y) = I^{i}(x,y) * \frac{1}{2\pi\sigma^2}e^{-\frac{x^2+y^2}{2\sigma^2}}$, where $\sigma$ is Gaussian standard deviation and $*$ denotes the convolution operator.
    \item \textit{Gamma correction}. We adjusted image gamma by a simple power-law expression: $I_\gamma^{i}(x,y) = I^{i}(x,y)^\gamma$, where $\gamma < 1$ compresses the gamma and $\gamma > 1$ results in gamma expansion.
    \item \textit{Salt \& Pepper noise}. This impulse noise is defined as: $I_{s\&p}^{i}(x,y) = I^i(x,y) + p_{s\&p}(z)\left\{
                \begin{array}{ll}
                  P_s \quad \textrm{for } z=s\\
                  P_p \quad \textrm{for } z=p\\
                  0 \quad \textrm{otherwise}
                \end{array}
              \right.$, where $p_{s\&p}$ is a probability density function (PDF).
    \item \textit{Gaussian noise}. This additive noise is defined as: $I_{\sigma_g}^{i}(x,y) = I^i(x,y) + p_{\sigma}(z)$, where $p_{\sigma}(z)$ is PDF of Gaussian distribution (\aka normal distribution) with variance $\sigma^2$ and zero mean ($\mu=0)$.
    \item \textit{Speckle noise}. This multiplicative noise is defined as: $I_{\sigma_s}^{i}(x,y) = I^i(x,y) \times (1 + p_{\sigma}(z))$, where $p_{\sigma}(z)$ is PDF of Gaussian distribution (\aka normal distribution) with variance $\sigma^2$ and zero mean ($\mu=0)$.
    \item \textit{Poisson noise}. Unlike all others that are independent of the image, Poisson noise is generated from pixel values, defined as: $I_\lambda^{i}(x,y) = p_\lambda^{i}(I(x,y))$, where $p_\lambda(z)$ is PDF of Poisson distribution defined as $\frac{\lambda^z e^{-\lambda}}{z!}$.
\end{enumerate}

It is worth emphasising that the human visual system exhibits a great amount of robustness to all these variations \cite{logothetis1996visual}, and therefore equally expected from machine vision. Moreover, from a practical point view, these conditions occur under typical circumstances, \eg, contrast and luminance alteration in natural scenes \cite{frazor2006local}, or presence of noise in images originated from various parts of the camera pipeline \cite{boncelet2009image}.

In the experiments reported in this article, we computed the classification accuracy of each network following these parameters\footnote{Source code and experimental materials are available at \url{https://goo.gl/eNpaUW}.}:
\begin{itemize}
    \item Contrast levels $c \in \left\{ 1, 5, 15, 30, 50, 75, 100 \right\} \%$.
    \item Illuminant ratios $l \in \left\{0.05, 0.25, 0.50, 0.75, 1.00 \right\}$.
    \item Gamma adjustment with $\gamma \in \left\{ 0.3, 0.8, 1.0, 1.2, 3.0 \right\}$.
    \item Image blurring with $\sigma \in \left\{0.0, 0.5, 1.0, 1.5 \right\}$.
    \item Percentage of independent noise $p \in \left\{0, 1, 5, 10 \right\} \%$.
\end{itemize}

\subsection{Models}

In this article, we restricted all our analysis to the family of \textit{ResNet} architecture \cite{he2016deep}. For \textit{CIFAR} data set we utilised \textit{ResNet20}, which consists of 274,442 parameters, and for \textit{ImageNet} a deeper version of it, namely \textit{ResNet50} that contains 25,636,712 parameters.

We trained 104 instances of \textit{ResNet20} on \textit{CIFAR-10}. Common configurations for all these networks include: 200 epochs on a single GPU, Adam optimiser \cite{kingma2014adam}, and following standard augmentation\footnote{We do not use the term augmentation in the sense of increasing the number of exposures. It merely refers to a manipulated version of the original image. Therefore, each network is essentially exposed to the same number of training images at each epoch.} procedures: \ie, random horizontal flipping, zooming (within a 10\% scale), and shifting (within a 10\% range). Parameters that were varied across different training procedures include: batch size, weights initialisation, learning rate, decay scheduler, and data augmentation through one or more of the image manipulations explained above.

We fine-tuned each of those 104 networks for 10 more epochs with the same set of configurations, except for the choice of image augmentation, for instance, some only with contrast reduction, others with additive noise, and a few with all or none. This procedure resulted in 225 fine-tuned networks. Therefore, overall we gathered 329 instances of \textit{ResNet20}.

We followed the same paradigm for \textit{ResNet50} on \textit{ImageNet}, although naturally with fewer instances due to its demanding computational resources. First, we trained 10 instances from scratch for 30 epochs, on a single GPU of batch size 32, randomly cropping images to 224 $\times$ 224 pixels. Note that during the evaluation this randomness was excluded and each image was resized to its smaller edge and the central square of side 224 was cropped.

Next, we picked one of the networks and fine-tuned it for 5 epochs to each individual of the eight image manipulations specified above. We further fine-tuned the same network on all image augmentations twice for 5 and 10 epochs. This procedure resulted in 10 fine-tuned versions. Therefore, overall we gathered 20 instances of \textit{ResNet50}.

\subsection{Network intrinsic similarity} 
\label{sec:nis}

There is no established technique to compute a measure of similarity between two neural networks. In this work, we defined the measure of similarity between a pair of networks (of an identical architecture) as the average Pearson correlation coefficient of their constituent kernels at every corresponding layer. Kernels are not positioned in an identical order within the same layer across different instances of a network. To account for this, we first aligned all kernels across the same layer of two networks in a one-to-one fashion according to their matching highest correlation coefficients. 

\section{Experiment and results}

\subsection{\textit{CIFAR}}

\begin{figure}[ht]
    \centering
    \includegraphics[width=1\columnwidth]{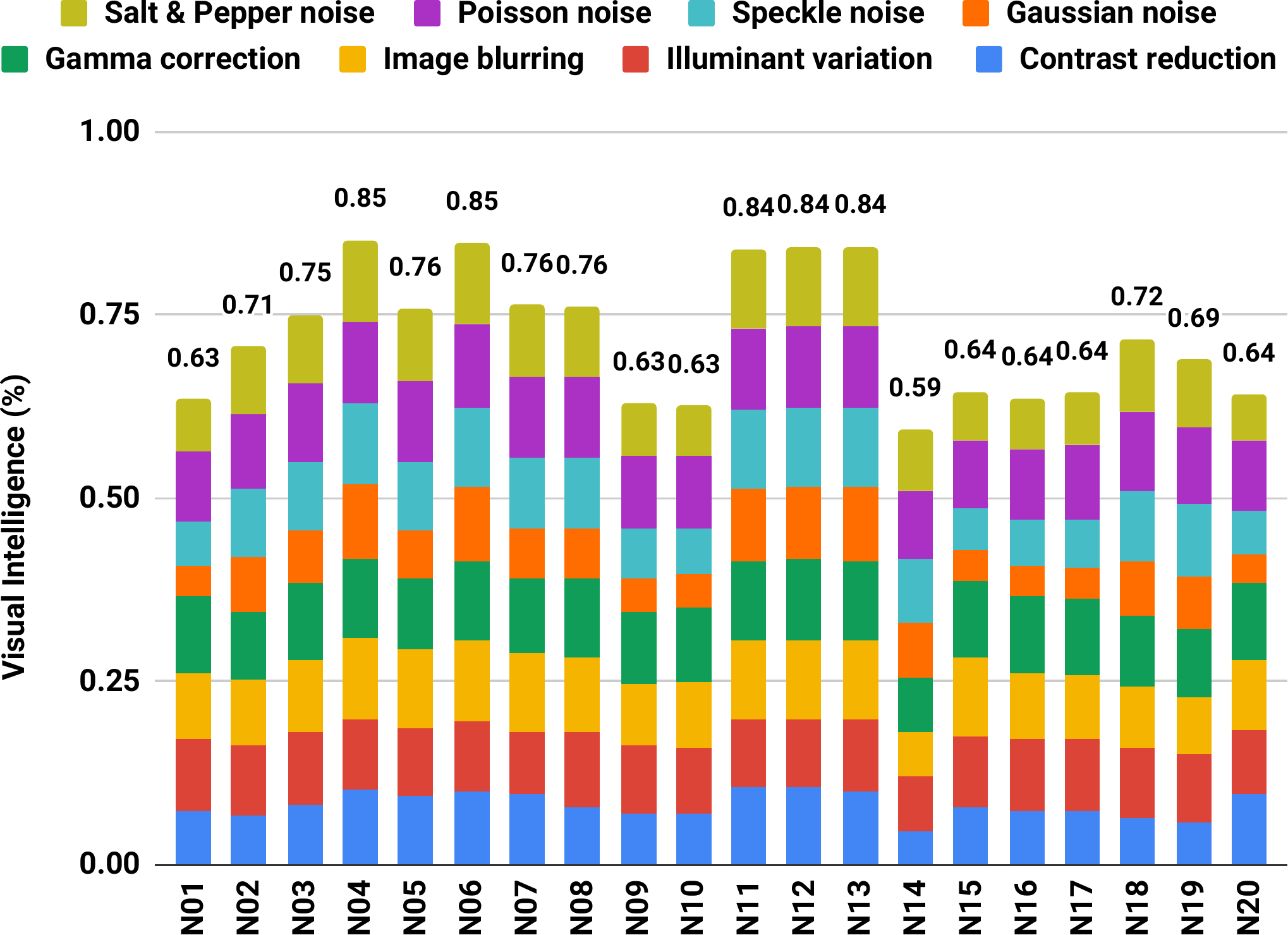}
    \caption{Distribution of classification accuracy across eight different tasks for a segment of studied networks (20 out of 329). The $y$-axis represents the average accuracy over eight types of image manipulation, which we refer to as \textbf{visual intelligence}.}
    \label{fig:20results}
\end{figure}

We computed a measure of performance (referred to as visual intelligence) for all instances of \textit{ResNet20} on the images of the \textit{CIFAR-10} validation set. The results for a representative subset of these networks (20 out of 329) is shown in Figure \ref{fig:20results}. Each bar represents a network and consists of eight segments, corresponding to the eight image manipulations defined in Section \ref{sec:congnition}. Classification accuracy of all these networks on the original images of the \textit{CIFAR-10} validation set without distortions is in the range of 0.88 to 0.92. With the distortions, their average performance has a larger range, between 0.59 to 0.85.

Evidently, various combinations could yield to a similar level of visual intelligence in its absolute term. However, this does not imply that the performance of networks is identical. For instance, \textit{N12} and \textit{N13} both score 0.84 in the measure of visual intelligence, but the former is more invariant to the changes of contrast and the latter to the illuminant. Therefore, we defined \textbf{visual intelligence compatibility} of a pair of networks as the Pearson correlation coefficient between their corresponding classification accuracy over all conducted experiments (refer to Figure \ref{fig:cifar10_vic_sup} in supplementary materials for a pairwise comparison of the same twenty networks).

\begin{figure*}[ht]
    \centering
    \includegraphics[width=1.6\columnwidth]{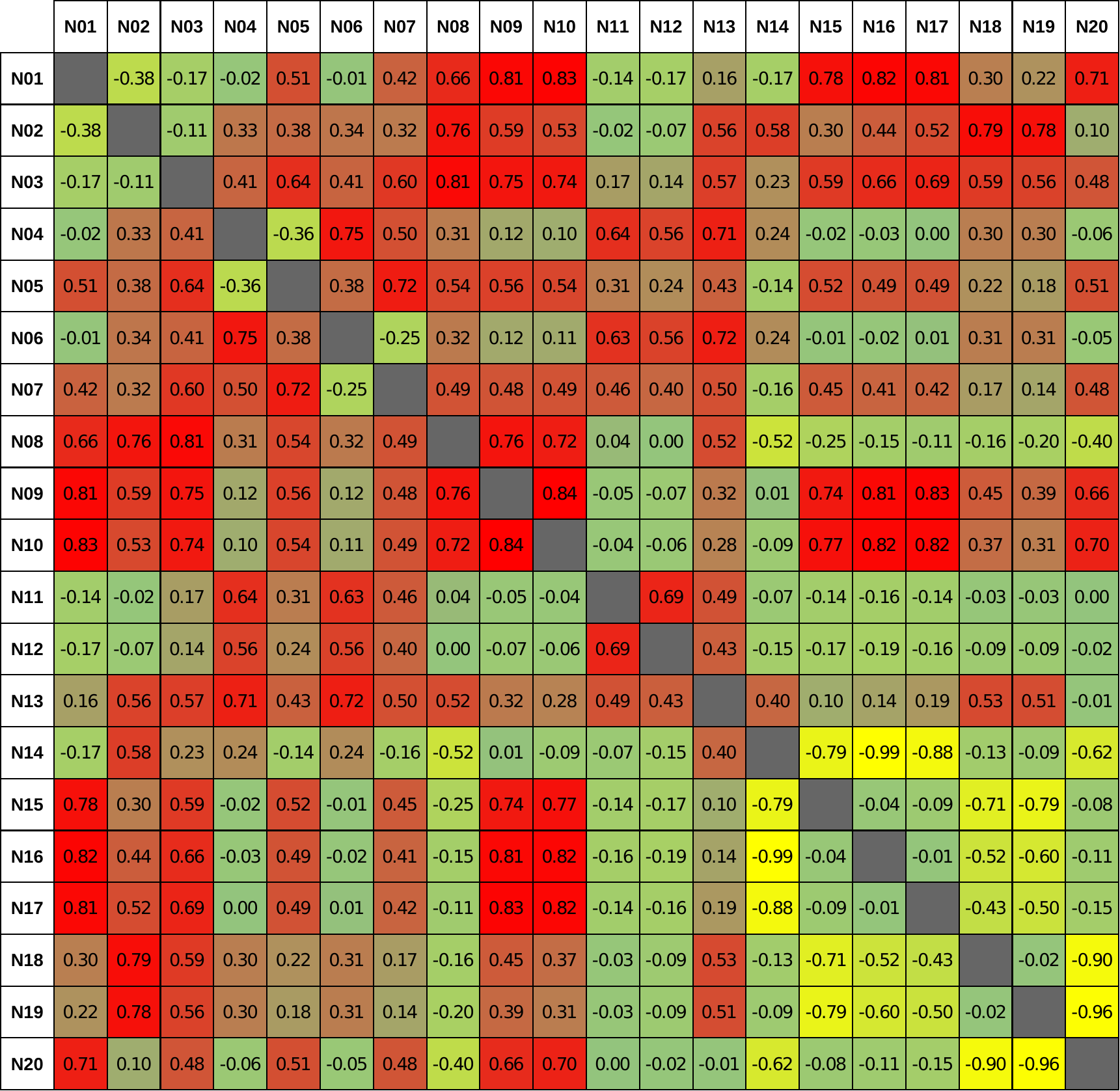}
    \caption{A pairwise comparison of twenty networks. Each cell represents the difference between two measures of visual intelligence compatibility and intrinsic networks similarity. Value \textbf{0} (\textbf{green} cells) indicates that both networks are identical and so is their visual intelligence. Value {\boldmath$-$}\textbf{1} (\textbf{yellow} cells) denotes kernels' weights are identical, yet the visual intelligence is different. Value \textbf{1} (\textbf{red} cells) signifies the opposite, two networks are distinct, yet their visual intelligence is identical.}
    \label{fig:cifar10_results}
\end{figure*}

Likewise, we computed a pairwise networks intrinsic similarity (as defined in Section \ref{sec:nis}) for all 53,956 pairs of \textit{ResNet20} (refer to Figure \ref{fig:cifar10_int_sup} in supplementary materials). Next, in order to investigate our hypothesis: whether there is an agreement between networks intrinsic similarity and their visual intelligence compatibility, we calculated the difference between these two measures. This comparison for the same set of twenty networks is reported in Figure \ref{fig:cifar10_results}.

If the intrinsic similarity of a pair of networks is on a par with their visual intelligence compatibility, the corresponding cell in Figure \ref{fig:cifar10_results} would be near to 0 (green cells). For instance, \textit{N16} and \textit{N17} are almost identical (over 99\% intrinsic similarity) and so is their performance across all manipulations (over 99\% visual intelligence compatibility). At the same time, \textit{N01} and \textit{N04} are only 20\% similar in both these measures. This agreement is rather expected.

If a pair of networks are intrinsically identical yet their visual intelligence is very different, their corresponding cell in Figure \ref{fig:cifar10_results} would be close to $-$1 (yellow cells). For instance, \textit{N14} and \textit{N16} are intrinsically the same (99\%), yet there is no compatibility in their visual intelligence (0\%), \ie, \textit{N16} obtains a better absolute performance, although \textit{N14} performs better in noisy images. Weight perturbations have been reported to influence the performance of DNN \cite{cheney2017robustness}, however not to the extent observed in our experiment. This is very puzzling on one side of the spectrum.

On the other side of the spectrum, if there is no intrinsic similarly between a pair of networks, yet their visual intelligence almost identical, the corresponding cell in Figure  \ref{fig:cifar10_results} would be nearly 1 (red cells). For instance, \textit{N01} and \textit{N10} are only 16\% intrinsically similar (expected from the distinct nature of image their training procedure), be that as it may, their visual intelligence is 99\% compatible. This is equally puzzling, that two distinct systems reach an indistinguishable performance across all eights manipulations. To emphasise, we do not refer merely to their absolute classification accuracy, rather a one-to-one correspondence in every detail of all distortions with many of them consisting random variables.

\subsection{\textit{ImageNet}}

\begin{figure}[!h]
    \centering
    \includegraphics[width=\columnwidth]{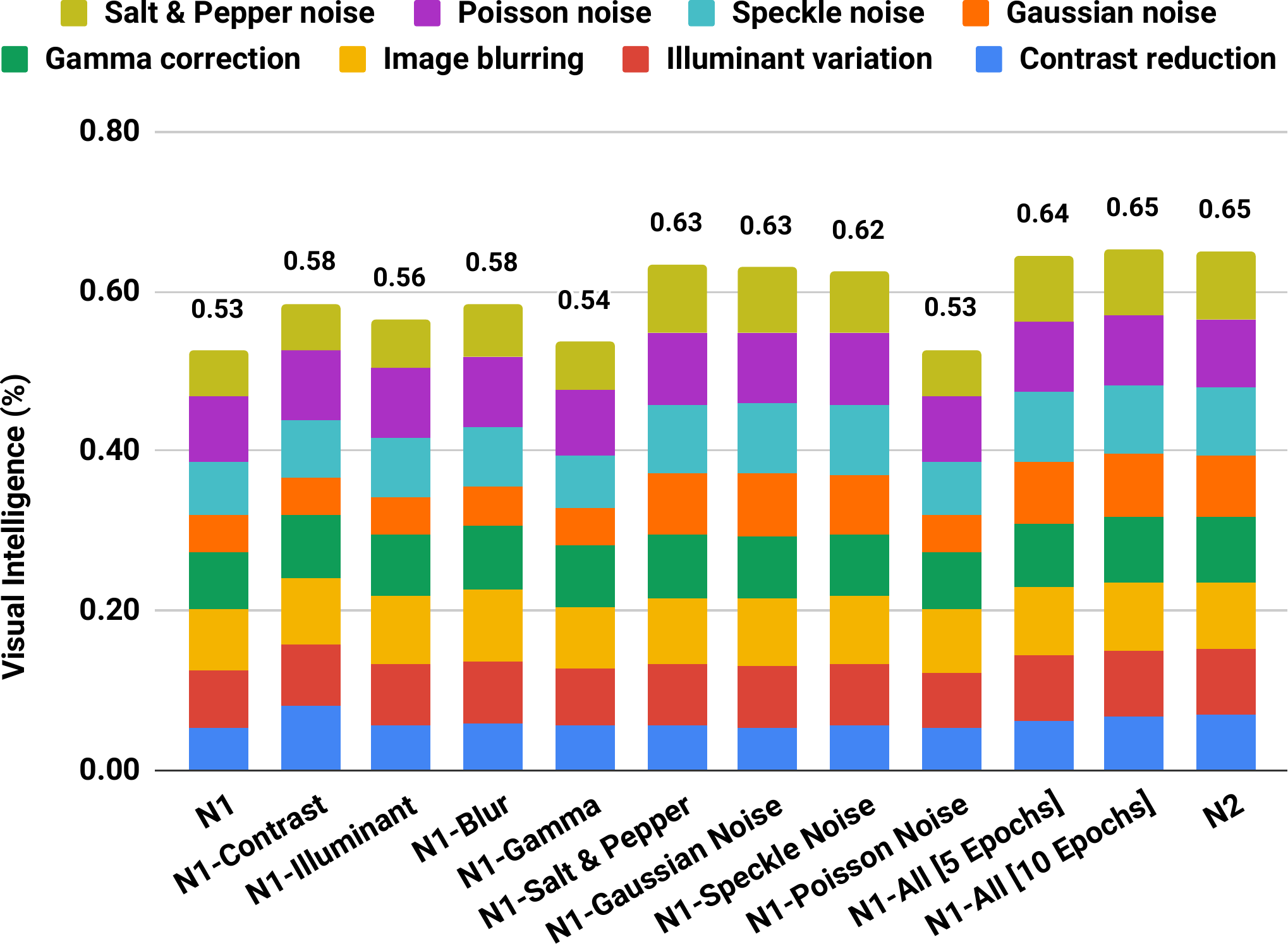}
    \caption{Distribution of classification accuracy across eight different tasks for twelve instances of \textit{ResNet50}. The $y$-axis represents the average accuracy over eight types of image manipulation, which we refer to as \textbf{visual intelligence}. \textit{N1} and \textit{N2} are trained from scratch, and all others are fine-tuned versions of \textit{N1}.}
    \label{fig:imagenet_bars}
\end{figure}

We conducted a similar procedure for all instances of \textit{ResNet50} on images of \textit{ImageNet} validation set. The visual intelligence of a subset of these networks is illustrated in Figure \ref{fig:imagenet_bars}. Networks \textit{N1} and \textit{N2} were trained from scratch with one difference: during the training procedure, \textit{N2} was exposed to those eight image manipulations and \textit{N1} was not. All others are fine-tuned versions of \textit{N1}, either with one distinct image augmentation (\eg, \textit{N1-Contrast} with images of a random contrast) or all the eight of them (\ie, \textit{N1-All}s). The classification accuracy of all obtained networks on original images of \textit{ImageNet} validation set with no distortion applied is in the range of 0.69 to 0.73. However, their average visual intelligence spans a larger spectrum, between 0.53 to 0.65.

\begin{figure*}[hb]
    \centering
    \hspace{2.2cm}
    \includegraphics[width=1.79\columnwidth]{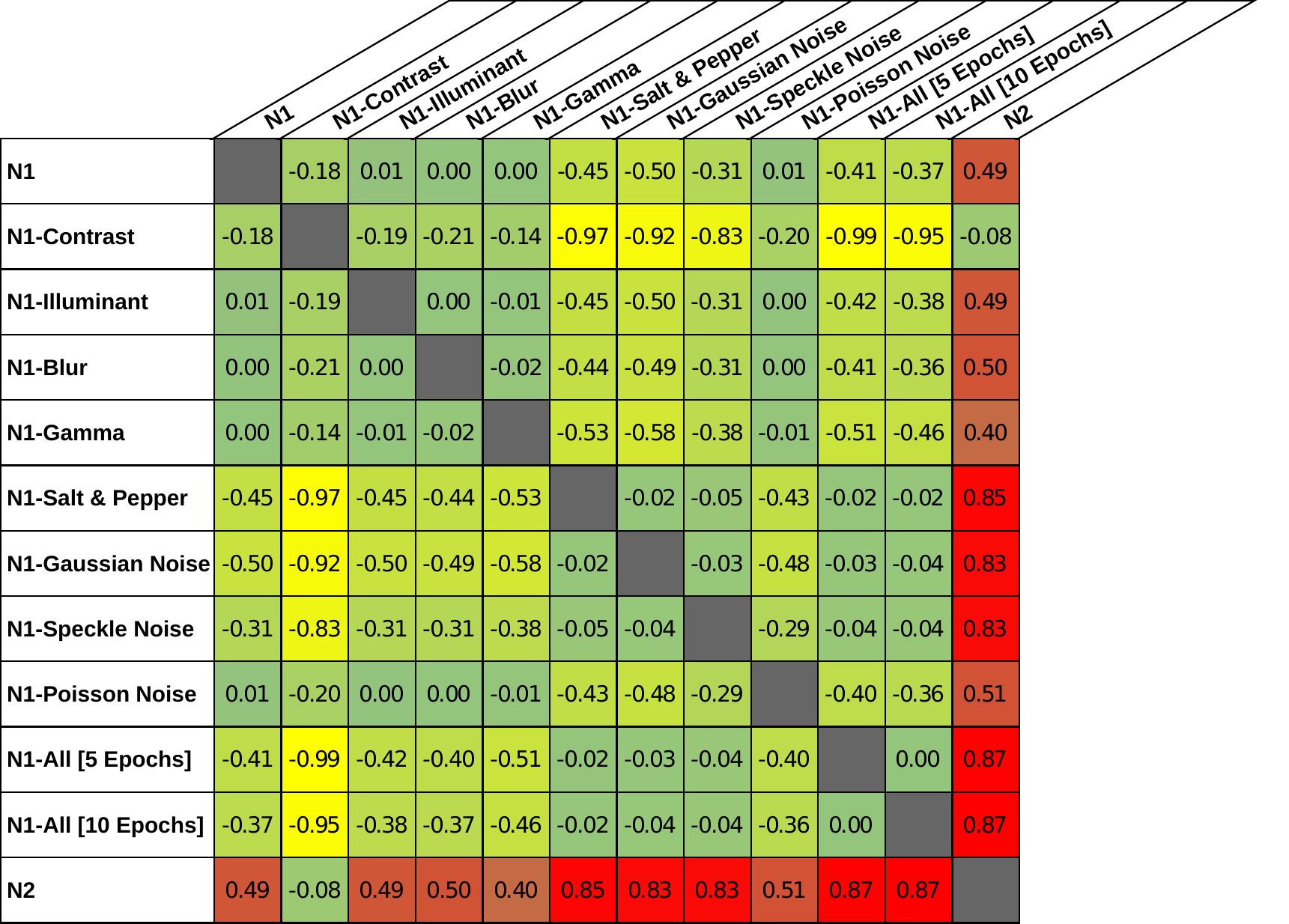}
    \caption{A pairwise comparison of twelve \textit{ResNet50}. Each cell represents the difference between two measures of visual intelligence compatibility and intrinsic networks similarity. Value \textbf{0} (\textbf{green} cells) indicates that both networks are identical and so is their visual intelligence. Value {\boldmath$-$}\textbf{1} (\textbf{yellow} cells) denotes kernels' weights are identical, yet the visual intelligence is different. Value \textbf{1} (\textbf{red} cells) signifies the opposite, two networks are distinct, yet their visual intelligence is identical.}
    \label{fig:imagenet_comparison}
\end{figure*}

It is worth highlighting that the original \textit{ResNet50} \cite{he2016deep} scores very similarly to \textit{N1} across all distortions. \textit{N1-All}s and \textit{N2} match their absolute performance on original images while outperforming them in all types of image manipulation. Although this is not a focus of our investigation, these results suggest that architectures similar to the original \textit{ResNet50} have the capacity to learn many more features without increasing their number of parameters \cite{akbarinia2019manifestation}, even to the level of surpassing human's performance for the same task of object classification under distorted images \cite{NIPS2018_7982}.

Correspondingly, we computed the visual intelligence compatibility and network intrinsic similarity for all 190 pairs of \textit{ResNet50} (refer to Figures \ref{fig:imagenet_vic_sup} and \ref{fig:imagenet_int_sup} in supplementary materials for a pairwise comparison of the same set twelve networks). The difference between these two measures is reported in Figure~\ref{fig:imagenet_comparison}. We can observe multiple instances of expected cases (green cells). For example, \textit{N1} and \textit{N1-Gamma} are identical intrinsically (99\%) and so is their visual intelligence compatibility (99\%). At the same time, comparison of many pairs result in this puzzling phenomenon. On the one hand, many of the \textit{N1}-offspring that are of the same nature, show no compatibility in their visual intelligence (yellow cells). On the other hand, \textit{N2} and \textit{N1-All}s that are intrinsically very different, reach a comparable level of visual intelligence (red cells).

\section{Discussion}

Comparison of deep neuronal networks (DNN) is often limited to their absolute performance, whether it is classification accuracy, computational time, or memory allocation, among others. Naturally, an infinite set of networks (both inter- and intra-architecture) could exhibit the same level of performance, irrespective of its type. The results of our experiments on image classification is another showcase of this fact. Several instances of \textit{ResNet} reached the same capacity of visual intelligence defined in this article (see Figures \ref{fig:20results} and \ref{fig:imagenet_bars}). Extending this form of performance comparison to constancy across multiple tasks, instead of its absolute value, does not alter the nature of this infinite set, even if it reduces it. For instance, our analysis demonstrated that various networks perform indistinguishably from each other under eight types of image distortion (refer to Figures \ref{fig:cifar10_vic_sup} and \ref{fig:imagenet_vic_sup} in supplementary materials).

As discussed in Section \ref{intro}, it is of great importance to understand the underlying mechanism of DNN. Arguably, one fundamental approach is to faithfully measure intrinsic similarity among these artificial networks (again, both inter- and intra-architecture). To the best of our knowledge, this remains an open question. The simple measure of cosine similarity has been reported to explain the intrinsic comparison in transfer learning in a very small dimensional space \cite{schneidermulti}. Here, we considered convolutional kernels as the smallest building block of a DNN and utilised Pearson correlation coefficient at this small dimension to capture intrinsic similarity of a pair of networks (refer to Figures \ref{fig:cifar10_int_sup} and \ref{fig:imagenet_int_sup} in supplementary materials).

We hypothesised that there should be a close correspondence between these two measures: the innate structure of a network and its cognitive ability (\ie, in our experiments, object detection in images under multiple conditions). This is not an unrealistic hypothesis. An analogy to parameters of a network is biological information---DNA and its four-based sequence. In this context, we share more cognitive abilities with other species of similar nature, \eg, apes. Humans have 96\% genomes in common with them \cite{sequencing2005initial}, many parts of our brain corresponds to theirs, \eg, visual cortex \cite{orban2004comparative}, and so does our ability in object detection \cite{rajalingham2015comparison}. Therefore, essentially we expected a reasonable matching between Figures \ref{fig:cifar10_vic_sup} and \ref{fig:cifar10_int_sup}, and Figures \ref{fig:imagenet_vic_sup} and \ref{fig:imagenet_int_sup} in supplementary materials.

However, the results of our experiments refute this hypothesis. We observed many cases in which intrinsic similarity of a pair of networks does not match to their classification performance (see Figures \ref{fig:cifar10_results} and \ref{fig:imagenet_comparison}). Those puzzling cases are not isolated and rather quite common. This could be due to several objections either to the choice of our method or the hypothesis itself.

\subsection{Research question}

The first objection would contend that the outcome of an optimisation procedure could give rise to an infinite set of solutions for the same problem, therefore, there are no grounds to believe that a correspondence between intrinsic similarity of networks and their performance should exist. We believe this claim is true to a certain extent. Two systems do not have to be identical to behave very similarly, as this is evident in the animal kingdom, although, there is a high probability that such systems would possess many characteristics in common. For instance, numerous aspects of the organisation of brain---a formidably complicated structural network---is well understood to maintain across inter- and intra-subjects \cite{bullmore2009complex}, so does in the hierarchy of a DNN \cite{lecun2015deep}. Therefore, the research question should hold true even if not in its absolute form, but to a large extent.

\subsection{Measure of intrinsic similarity}

The second objection would be that the Pearson correlation coefficient does not capture the complexity of a DNN. Certainly, correlation alone cannot account for the complexity of a network, in which structure and function are twisted together. Therefore, a comprehensive measure of intrinsic similarity between a pair of networks requires a topological approach \cite{borner2007network}. Nevertheless, there are reasons to believe that this simple measure should at least account for a part of the similarity at the level of kernels.
\begin{enumerate}
    \item This notion that kernels of the first layer resemble Gabor filters or colour blobs \cite{yosinski2014transferable}, implies that individual kernels can be considered as smallest building blocks of a complicated structural network, tuned to a specific stimulus, and consequently there should be a sort of similarity between them across networks. Furthermore, it has been reported that overall architecture of a DNN is reminiscent of the LGN--V1--V2--V4--IT hierarchy in the visual cortex \cite{lecun2015deep}, this, in turn, strengthens a layer-to-layer comparison of a pair of networks given their constituent kernels are faithfully aligned. Indeed, it has been reported \cite{Bau_2017_CVPR} that training conditions does not affect dissection of \textit{AlexNet} \cite{krizhevsky2012imagenet} into interpretable concepts. Given this finding is grounded on kernels' activation maps, it can be argued that their underlying kernels' parameters should perform very similar operations.
    \item We computed the correlation coefficient for a pair of parent-child networks (see Figure \ref{fig:resnet50_layers}). This indicates in which layers the child network has deviated more from its parent (\eg, here the convolutional layer \#20, \ie, \textit{res3c\_branch2c} in \textit{ResNet50}). Next, we directly transferred (literally copying) weights of \textit{res3c\_branch2c} from \textit{N1-All} (child) to \textit{N1} (parent). The new network reaches the visual intelligent score of 0.55, \ie, 2\% more than the original \textit{N1}. This is a significant improvement, considering two facts: (i) \textit{res3c\_branch2c} contains only 66,048 parameters, which is a tiny fraction of the entire model (about 0.2\%), and (ii) all its inputs and outputs connections are not changed. Correspondingly, transferring weights of all convolutional layers from \textit{N1-All} to \textit{N1} produces a network with the same level of performance as \textit{N1-All}. Again, this is very significant, considering the fully connected layer, which accounts for 8\% of network's parameters, remains intact.
    \begin{figure}[ht]
        \centering
        \includegraphics[width=1\columnwidth]{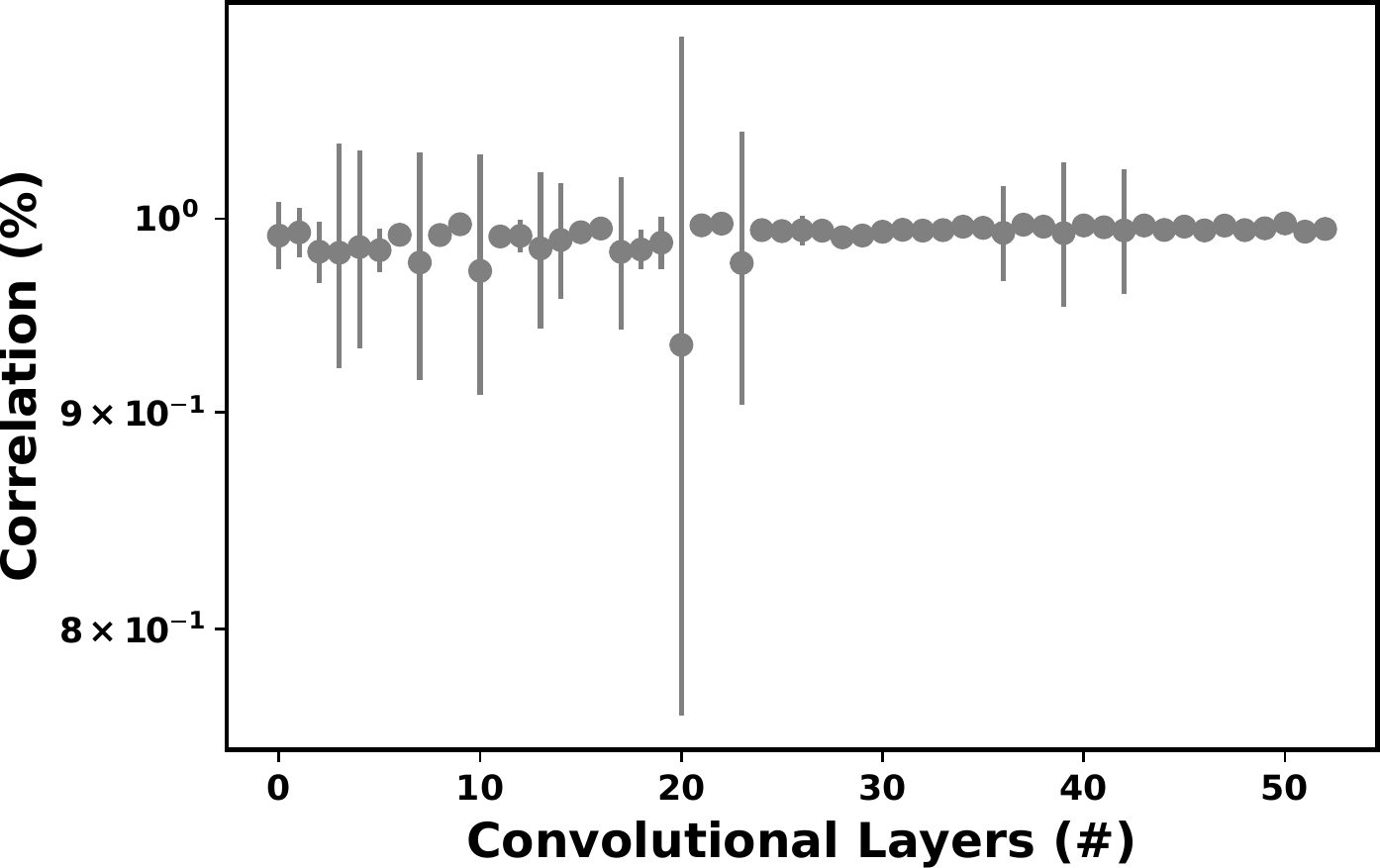}
        \caption{The Pearson correlation coefficient between all convolutional layers of two \textit{ResNet50} networks---\textit{N1} and its fine-tuned offspring \textit{N1-All}. Each data point denotes the average of all kernels for that layer. Vertical lines represent $\pm1$ standard deviation.}
        \label{fig:resnet50_layers}
    \end{figure}
\end{enumerate}

\subsection{Visual tasks}

The third objection would state that the set of conducted evaluations to define the visual intelligence of networks is not well grounded. Perhaps a wider range of tasks is necessary to truly capture differences among networks. Naturally, the larger the information space, the more likelihood of differentiating, however, we believe image manipulations defined in our experiments include challenging realistic scenarios that are necessary for robust object detection. Furthermore, the evaluation phase included conditions unseen during the training, which has been shown to greatly trouble generalisation of DNNs \cite{NIPS2018_7982}. Therefore, the visual intelligence compatibility computed from each pair of networks cannot be a coincidence and captures similarity of cognitive ability to a reasonable level.

Recently, it has been demonstrated that an intertwined feature representation emerges for a network trained on multiple high-level cognitive tasks \cite{task2019}. Our approach differs in two aspects: (i) the variation in our tasks are of a low-level nature and (ii) we analysed kernels' weights instead of their response. Nevertheless, this compositionality of task representations is present in our experiments. The convolutional layer \textit{res3c\_branch2c}, discussed above, exhibits the same characteristic as Figure \ref{fig:resnet50_layers} for all other offspring of \textit{N1}. This means the very same layer encodes information about multiple low-level visual feature, such as the contrast of input image or presence of noise. Furthermore, the direct transfer of this layer's weights did not only boost the absolute visual intelligence by 2\%, but also it lead to an improvement in every single task, reinforcing this idea that various low-level features are integrated into that layer.

\subsection{Future works}

Most works in the literature seek feature representation from activity of kernels exposed to  high-level concepts, \eg, \cite{Bau_2017_CVPR,task2019}. Although their findings are of great importance for the interpretation of DNNs, their general assumption is based on a latent variable perspective according to the hierarchical representation of of data across layers. Recently, a novel perspective has been proposed to approach kernels operations with Riemannian geometry, in which consecutive layers successively warp the coordinate representation of training data \cite{hauser2017principles}. In the conducted comparisons between \textit{N1} and its fine-tuned offspring, we observed that variation in low-level features (\eg, contrast, illuminant, and noise) highly influenced a middle layer (\textit{res3c\_branch2c} in \textit{ResNet50}). Perhaps, subtle changes in preceding layers has lead to a complex representation of data manifold at this layer. Therefore, we propose a geometrical matching of kernels (with their raw weights) to measure intrinsic similarity of two networks.

\section{Conclusion}

Throughout this article, we argued for a correspondence between the intrinsic similarity of networks and their performance. The results of our experiments come short of demonstrating this hypothesis. This mainly raises awareness of a need to a more adequate comparison of kernels' weights and their connections between a pair of networks. Future works on this line of investigation would allow us to better quantify similarities and differences among various deep neuronal networks, which, in turn, would be a step forward towards a more profound understanding of their intelligent and hopefully our own.

\subsubsection*{Acknowledgements}

This project was funded by the Deutsche Forschungsgemeinschaft SFB/TRR 135.

\bibliography{ICML19_187}

\begin{thebibliography}{44}
\providecommand{\natexlab}[1]{#1}
\providecommand{\url}[1]{\texttt{#1}}
\expandafter\ifx\csname urlstyle\endcsname\relax
  \providecommand{\doi}[1]{doi: #1}\else
  \providecommand{\doi}{doi: \begingroup \urlstyle{rm}\Url}\fi

\bibitem[Agrawal et~al.(2014)Agrawal, Girshick, and
  Malik]{agrawal2014analyzing}
Agrawal, P., Girshick, R., and Malik, J.
\newblock Analyzing the performance of multilayer neural networks for object
  recognition.
\newblock In \emph{European conference on computer vision}, pp.\  329--344.
  Springer, 2014.

\bibitem[Akbarinia \& Gegenfurtner(2018)Akbarinia and
  Gegenfurtner]{akbarinia2018contrast}
Akbarinia, A. and Gegenfurtner, K.~R.
\newblock How is contrast encoded in deep neural networks?
\newblock \emph{arXiv preprint arXiv:1809.01438}, 2018.

\bibitem[Akbarinia \& Gegenfurtner(2019)Akbarinia and
  Gegenfurtner]{akbarinia2019manifestation}
Akbarinia, A. and Gegenfurtner, K.~R.
\newblock Manifestation of image contrast in deep networks.
\newblock \emph{arXiv preprint arXiv:1902.04378}, 2019.

\bibitem[Bau et~al.(2017)Bau, Zhou, Khosla, Oliva, and Torralba]{Bau_2017_CVPR}
Bau, D., Zhou, B., Khosla, A., Oliva, A., and Torralba, A.
\newblock Network dissection: Quantifying interpretability of deep visual
  representations.
\newblock In \emph{The IEEE Conference on Computer Vision and Pattern
  Recognition (CVPR)}, July 2017.

\bibitem[Bengio et~al.(2013)Bengio, Courville, and
  Vincent]{bengio2013representation}
Bengio, Y., Courville, A., and Vincent, P.
\newblock Representation learning: A review and new perspectives.
\newblock \emph{IEEE transactions on pattern analysis and machine
  intelligence}, 35\penalty0 (8):\penalty0 1798--1828, 2013.

\bibitem[Boncelet(2009)]{boncelet2009image}
Boncelet, C.
\newblock Image noise models.
\newblock In \emph{The Essential Guide to Image Processing}, pp.\  143--167.
  Elsevier, 2009.

\bibitem[B{\"o}rner et~al.(2007)B{\"o}rner, Sanyal, and
  Vespignani]{borner2007network}
B{\"o}rner, K., Sanyal, S., and Vespignani, A.
\newblock Network science.
\newblock \emph{Annual review of information science and technology},
  41\penalty0 (1):\penalty0 537--607, 2007.

\bibitem[Bullmore \& Sporns(2009)Bullmore and Sporns]{bullmore2009complex}
Bullmore, E. and Sporns, O.
\newblock Complex brain networks: graph theoretical analysis of structural and
  functional systems.
\newblock \emph{Nature Reviews Neuroscience}, 10\penalty0 (3):\penalty0 186,
  2009.

\bibitem[Cheney et~al.(2017)Cheney, Schrimpf, and
  Kreiman]{cheney2017robustness}
Cheney, N., Schrimpf, M., and Kreiman, G.
\newblock On the robustness of convolutional neural networks to internal
  architecture and weight perturbations.
\newblock \emph{arXiv preprint arXiv:1703.08245}, 2017.

\bibitem[Eckstein et~al.(2017)Eckstein, Koehler, Welbourne, and
  Akbas]{eckstein2017humans}
Eckstein, M.~P., Koehler, K., Welbourne, L.~E., and Akbas, E.
\newblock Humans, but not deep neural networks, often miss giant targets in
  scenes.
\newblock \emph{Current Biology}, 27\penalty0 (18):\penalty0 2827--2832, 2017.

\bibitem[Flachot \& Gegenfurtner(2018)Flachot and
  Gegenfurtner]{flachot2018processing}
Flachot, A. and Gegenfurtner, K.~R.
\newblock Processing of chromatic information in a deep convolutional neural
  network.
\newblock \emph{JOSA A}, 35\penalty0 (4):\penalty0 B334--B346, 2018.

\bibitem[Frazor \& Geisler(2006)Frazor and Geisler]{frazor2006local}
Frazor, R.~A. and Geisler, W.~S.
\newblock Local luminance and contrast in natural images.
\newblock \emph{Vision research}, 46\penalty0 (10):\penalty0 1585--1598, 2006.

\bibitem[Geirhos et~al.(2018)Geirhos, Temme, Rauber, Sch\"{u}tt, Bethge, and
  Wichmann]{NIPS2018_7982}
Geirhos, R., Temme, C. R.~M., Rauber, J., Sch\"{u}tt, H.~H., Bethge, M., and
  Wichmann, F.~A.
\newblock Generalisation in humans and deep neural networks.
\newblock In Bengio, S., Wallach, H., Larochelle, H., Grauman, K.,
  Cesa-Bianchi, N., and Garnett, R. (eds.), \emph{Advances in Neural
  Information Processing Systems 31}, pp.\  7549--7561. Curran Associates,
  Inc., 2018.

\bibitem[Glorot \& Bengio(2010)Glorot and Bengio]{glorot2010understanding}
Glorot, X. and Bengio, Y.
\newblock Understanding the difficulty of training deep feedforward neural
  networks.
\newblock In \emph{Proceedings of the thirteenth international conference on
  artificial intelligence and statistics}, pp.\  249--256, 2010.

\bibitem[Hauser \& Ray(2017)Hauser and Ray]{hauser2017principles}
Hauser, M. and Ray, A.
\newblock Principles of riemannian geometry in neural networks.
\newblock In \emph{Advances in Neural Information Processing Systems}, pp.\
  2807--2816, 2017.

\bibitem[He et~al.(2016)He, Zhang, Ren, and Sun]{he2016deep}
He, K., Zhang, X., Ren, S., and Sun, J.
\newblock Deep residual learning for image recognition.
\newblock In \emph{Proceedings of the IEEE conference on computer vision and
  pattern recognition}, pp.\  770--778, 2016.

\bibitem[Kell et~al.(2018)Kell, Yamins, Shook, Norman-Haignere, and
  McDermott]{kell2018task}
Kell, A.~J., Yamins, D.~L., Shook, E.~N., Norman-Haignere, S.~V., and
  McDermott, J.~H.
\newblock A task-optimized neural network replicates human auditory behavior,
  predicts brain responses, and reveals a cortical processing hierarchy.
\newblock \emph{Neuron}, 98\penalty0 (3):\penalty0 630--644, 2018.

\bibitem[Kingma \& Ba(2014)Kingma and Ba]{kingma2014adam}
Kingma, D.~P. and Ba, J.
\newblock Adam: A method for stochastic optimization.
\newblock \emph{arXiv preprint arXiv:1412.6980}, 2014.

\bibitem[Krizhevsky \& Hinton(2009)Krizhevsky and
  Hinton]{krizhevsky2009learning}
Krizhevsky, A. and Hinton, G.
\newblock Learning multiple layers of features from tiny images.
\newblock Technical report, Citeseer, 2009.

\bibitem[Krizhevsky et~al.(2012)Krizhevsky, Sutskever, and
  Hinton]{krizhevsky2012imagenet}
Krizhevsky, A., Sutskever, I., and Hinton, G.~E.
\newblock Imagenet classification with deep convolutional neural networks.
\newblock In \emph{Advances in neural information processing systems}, pp.\
  1097--1105, 2012.

\bibitem[LeCun et~al.(2015)LeCun, Bengio, and Hinton]{lecun2015deep}
LeCun, Y., Bengio, Y., and Hinton, G.
\newblock Deep learning.
\newblock \emph{nature}, 521\penalty0 (7553):\penalty0 436, 2015.

\bibitem[Logothetis \& Sheinberg(1996)Logothetis and
  Sheinberg]{logothetis1996visual}
Logothetis, N.~K. and Sheinberg, D.~L.
\newblock Visual object recognition.
\newblock \emph{Annual review of neuroscience}, 19\penalty0 (1):\penalty0
  577--621, 1996.

\bibitem[Mahendran \& Vedaldi(2015)Mahendran and
  Vedaldi]{mahendran2015understanding}
Mahendran, A. and Vedaldi, A.
\newblock Understanding deep image representations by inverting them.
\newblock In \emph{Proceedings of the IEEE conference on computer vision and
  pattern recognition}, pp.\  5188--5196, 2015.

\bibitem[Marblestone et~al.(2016)Marblestone, Wayne, and
  Kording]{marblestone2016toward}
Marblestone, A.~H., Wayne, G., and Kording, K.~P.
\newblock Toward an integration of deep learning and neuroscience.
\newblock \emph{Frontiers in computational neuroscience}, 10:\penalty0 94,
  2016.

\bibitem[Moosavi-Dezfooli et~al.(2016)Moosavi-Dezfooli, Fawzi, and
  Frossard]{moosavi2016deepfool}
Moosavi-Dezfooli, S.-M., Fawzi, A., and Frossard, P.
\newblock Deepfool: a simple and accurate method to fool deep neural networks.
\newblock In \emph{Proceedings of the IEEE Conference on Computer Vision and
  Pattern Recognition}, pp.\  2574--2582, 2016.

\bibitem[Orban et~al.(2004)Orban, Van~Essen, and
  Vanduffel]{orban2004comparative}
Orban, G.~A., Van~Essen, D., and Vanduffel, W.
\newblock Comparative mapping of higher visual areas in monkeys and humans.
\newblock \emph{Trends in cognitive sciences}, 8\penalty0 (7):\penalty0
  315--324, 2004.

\bibitem[Papernot et~al.(2017)Papernot, McDaniel, Goodfellow, Jha, Celik, and
  Swami]{papernot2017practical}
Papernot, N., McDaniel, P., Goodfellow, I., Jha, S., Celik, Z.~B., and Swami,
  A.
\newblock Practical black-box attacks against machine learning.
\newblock In \emph{Proceedings of the 2017 ACM on Asia Conference on Computer
  and Communications Security}, pp.\  506--519. ACM, 2017.

\bibitem[Pryluk et~al.(2019)Pryluk, Kfir, Gelbard-Sagiv, Fried, and
  Paz]{pryluk2019tradeoff}
Pryluk, R., Kfir, Y., Gelbard-Sagiv, H., Fried, I., and Paz, R.
\newblock A tradeoff in the neural code across regions and species.
\newblock \emph{Cell}, 2019.

\bibitem[Quiroga et~al.(2005)Quiroga, Reddy, Kreiman, Koch, and
  Fried]{quiroga2005invariant}
Quiroga, R.~Q., Reddy, L., Kreiman, G., Koch, C., and Fried, I.
\newblock Invariant visual representation by single neurons in the human brain.
\newblock \emph{Nature}, 435\penalty0 (7045):\penalty0 1102, 2005.

\bibitem[Rajalingham et~al.(2015)Rajalingham, Schmidt, and
  DiCarlo]{rajalingham2015comparison}
Rajalingham, R., Schmidt, K., and DiCarlo, J.~J.
\newblock Comparison of object recognition behavior in human and monkey.
\newblock \emph{Journal of Neuroscience}, 35\penalty0 (35):\penalty0
  12127--12136, 2015.

\bibitem[Schneider et~al.(2018)Schneider, Ecker, Macke, and
  Bethge]{schneidermulti}
Schneider, S., Ecker, A.~S., Macke, J.~H., and Bethge, M.
\newblock Multi-task generalization and adaptation between noisy digit
  datasets: An empirical study.
\newblock \emph{1Neural Information Processing Systems (NIPS) Workshop on
  Continual Learning}, 2018.

\bibitem[Sequencing et~al.(2005)Sequencing, Waterson, Lander, Wilson,
  Consortium, et~al.]{sequencing2005initial}
Sequencing, T.~C., Waterson, R.~H., Lander, E.~S., Wilson, R.~K., Consortium,
  A., et~al.
\newblock Initial sequence of the chimpanzee genome and comparison with the
  human genome.
\newblock \emph{Nature}, 437\penalty0 (7055):\penalty0 69, 2005.

\bibitem[Silver et~al.(2018)Silver, Hubert, Schrittwieser, Antonoglou, Lai,
  Guez, Lanctot, Sifre, Kumaran, Graepel, et~al.]{silver2018general}
Silver, D., Hubert, T., Schrittwieser, J., Antonoglou, I., Lai, M., Guez, A.,
  Lanctot, M., Sifre, L., Kumaran, D., Graepel, T., et~al.
\newblock A general reinforcement learning algorithm that masters chess, shogi,
  and go through self-play.
\newblock \emph{Science}, 362\penalty0 (6419):\penalty0 1140--1144, 2018.

\bibitem[Simonyan et~al.(2013)Simonyan, Vedaldi, and
  Zisserman]{simonyan2013deep}
Simonyan, K., Vedaldi, A., and Zisserman, A.
\newblock Deep inside convolutional networks: Visualising image classification
  models and saliency maps.
\newblock \emph{arXiv preprint arXiv:1312.6034}, 2013.

\bibitem[Stallkamp et~al.(2012)Stallkamp, Schlipsing, Salmen, and
  Igel]{stallkamp2012man}
Stallkamp, J., Schlipsing, M., Salmen, J., and Igel, C.
\newblock Man vs. computer: Benchmarking machine learning algorithms for
  traffic sign recognition.
\newblock \emph{Neural networks}, 32:\penalty0 323--332, 2012.

\bibitem[Szegedy et~al.(2013)Szegedy, Zaremba, Sutskever, Bruna, Erhan,
  Goodfellow, and Fergus]{szegedy2013intriguing}
Szegedy, C., Zaremba, W., Sutskever, I., Bruna, J., Erhan, D., Goodfellow, I.,
  and Fergus, R.
\newblock Intriguing properties of neural networks.
\newblock \emph{arXiv preprint arXiv:1312.6199}, 2013.

\bibitem[Tian et~al.(2018)Tian, Pei, Jana, and Ray]{tian2018deeptest}
Tian, Y., Pei, K., Jana, S., and Ray, B.
\newblock Deeptest: Automated testing of deep-neural-network-driven autonomous
  cars.
\newblock In \emph{Proceedings of the 40th International Conference on Software
  Engineering}, pp.\  303--314. ACM, 2018.

\bibitem[Titano et~al.(2018)Titano, Badgeley, Schefflein, Pain, Su, Cai,
  Swinburne, Zech, Kim, Bederson, et~al.]{titano2018automated}
Titano, J.~J., Badgeley, M., Schefflein, J., Pain, M., Su, A., Cai, M.,
  Swinburne, N., Zech, J., Kim, J., Bederson, J., et~al.
\newblock Automated deep-neural-network surveillance of cranial images for
  acute neurologic events.
\newblock \emph{Nat Med}, 24\penalty0 (9):\penalty0 1337--1341, 2018.

\bibitem[Yang et~al.(2019)Yang, Joglekar, Song, Newsome, and Wang]{task2019}
Yang, G.~R., Joglekar, M.~R., Song, H.~F., Newsome, W.~T., and Wang, X.-J.
\newblock Task representations in neural networks trained to perform many
  cognitive tasks.
\newblock \emph{Nature Neuroscience}, 1\penalty0 (1):\penalty0 1546--1726,
  2019.

\bibitem[Yosinski et~al.(2014)Yosinski, Clune, Bengio, and
  Lipson]{yosinski2014transferable}
Yosinski, J., Clune, J., Bengio, Y., and Lipson, H.
\newblock How transferable are features in deep neural networks?
\newblock In \emph{Advances in neural information processing systems}, pp.\
  3320--3328, 2014.

\bibitem[Zeiler \& Fergus(2014)Zeiler and Fergus]{zeiler2014visualizing}
Zeiler, M.~D. and Fergus, R.
\newblock Visualizing and understanding convolutional networks.
\newblock In \emph{European conference on computer vision}, pp.\  818--833.
  Springer, 2014.

\bibitem[Zhang et~al.(2018)Zhang, Wu, and Zhu]{zhang2018interpretable}
Zhang, Q., Wu, Y.~N., and Zhu, S.-C.
\newblock Interpretable convolutional neural networks.
\newblock In \emph{The IEEE Conference on Computer Vision and Pattern
  Recognition (CVPR)}, pp.\  8827--8836, 2018.

\bibitem[Zhang \& Zhu(2018)Zhang and Zhu]{zhang2018visual}
Zhang, Q.-s. and Zhu, S.-C.
\newblock Visual interpretability for deep learning: a survey.
\newblock \emph{Frontiers of Information Technology \& Electronic Engineering},
  19\penalty0 (1):\penalty0 27--39, 2018.

\bibitem[Zhao et~al.(2018)Zhao, Daley, and Pruszynski]{zhao2018neural}
Zhao, C.~W., Daley, M.~J., and Pruszynski, J.~A.
\newblock Neural network models of the tactile system develop first-order units
  with spatially complex receptive fields.
\newblock \emph{PloS one}, 13\penalty0 (6):\penalty0 e0199196, 2018.

\end{thebibliography}
\bibliographystyle{icml2019}

\appendix

\section{\textit{CIFAR-10}}
\begin{figure*}[ht]
    \centering
    \includegraphics[width=2\columnwidth]{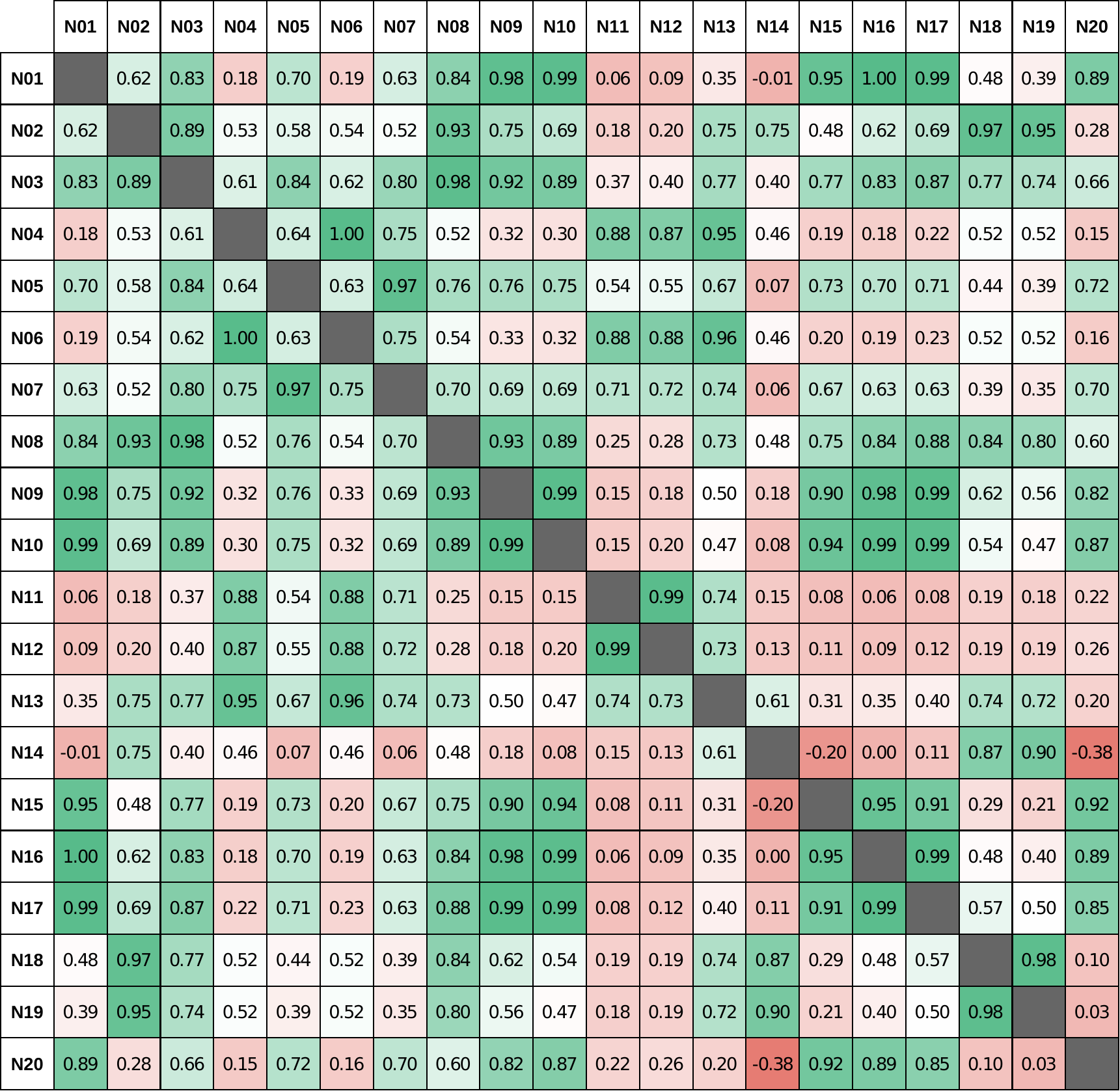}
    \caption{Pairwise comparison of visual intelligence compatibility between twenty networks. All networks are of \textit{ResNet20} architecture and have been trained on \textit{CIFAR-10} dataset. \textbf{Green} cells indicate high compatibility between performance of a pair of networks across all eight image distortions. \textbf{Red} cells denote the opposite, no correlation between the performance of two networks.}
    \label{fig:cifar10_vic_sup}
\end{figure*}

Figure \ref{fig:cifar10_vic_sup} illustrates visual intelligence compatibility of a subset of studied \textit{ResNet20}.

\begin{figure*}[ht]
    \centering
    \includegraphics[width=2\columnwidth]{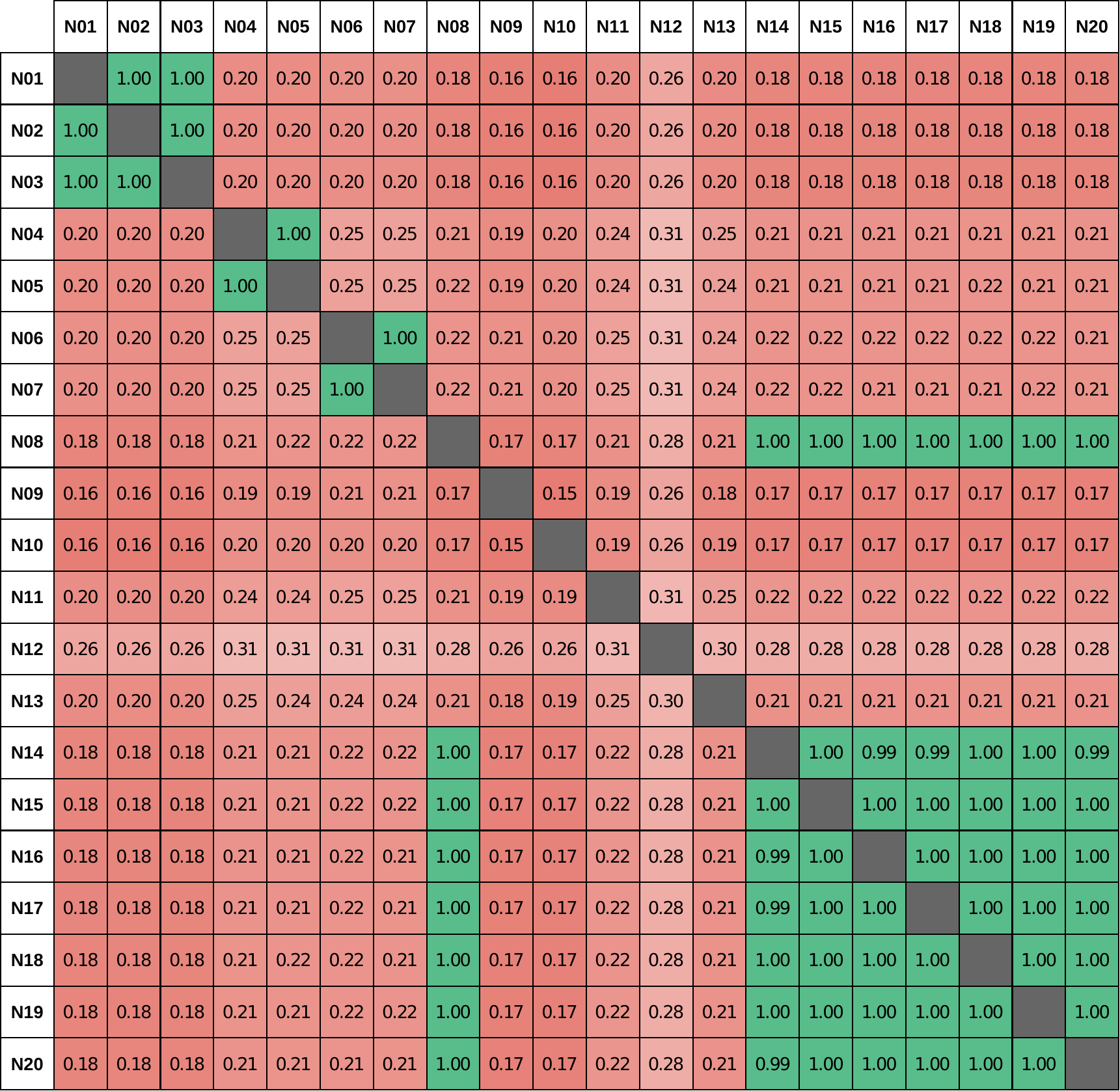}
    \caption{Pairwise comparison of intrinsic similarity between twenty networks. All networks are of \textit{ResNet20} architecture and have been trained on \textit{CIFAR-10} dataset. \textbf{Green} cells indicate high Pearson correlation coefficient between kernels' weights of a pair of networks. \textbf{Red} cells denote the opposite, no correlation between kernels' weights of two networks.}
    \label{fig:cifar10_int_sup}
\end{figure*}

Figure \ref{fig:cifar10_int_sup} illustrates intrinsic similarity of a subset of studied \textit{ResNet20}.

\begin{figure*}[ht]
    \centering
    \includegraphics[width=2\columnwidth]{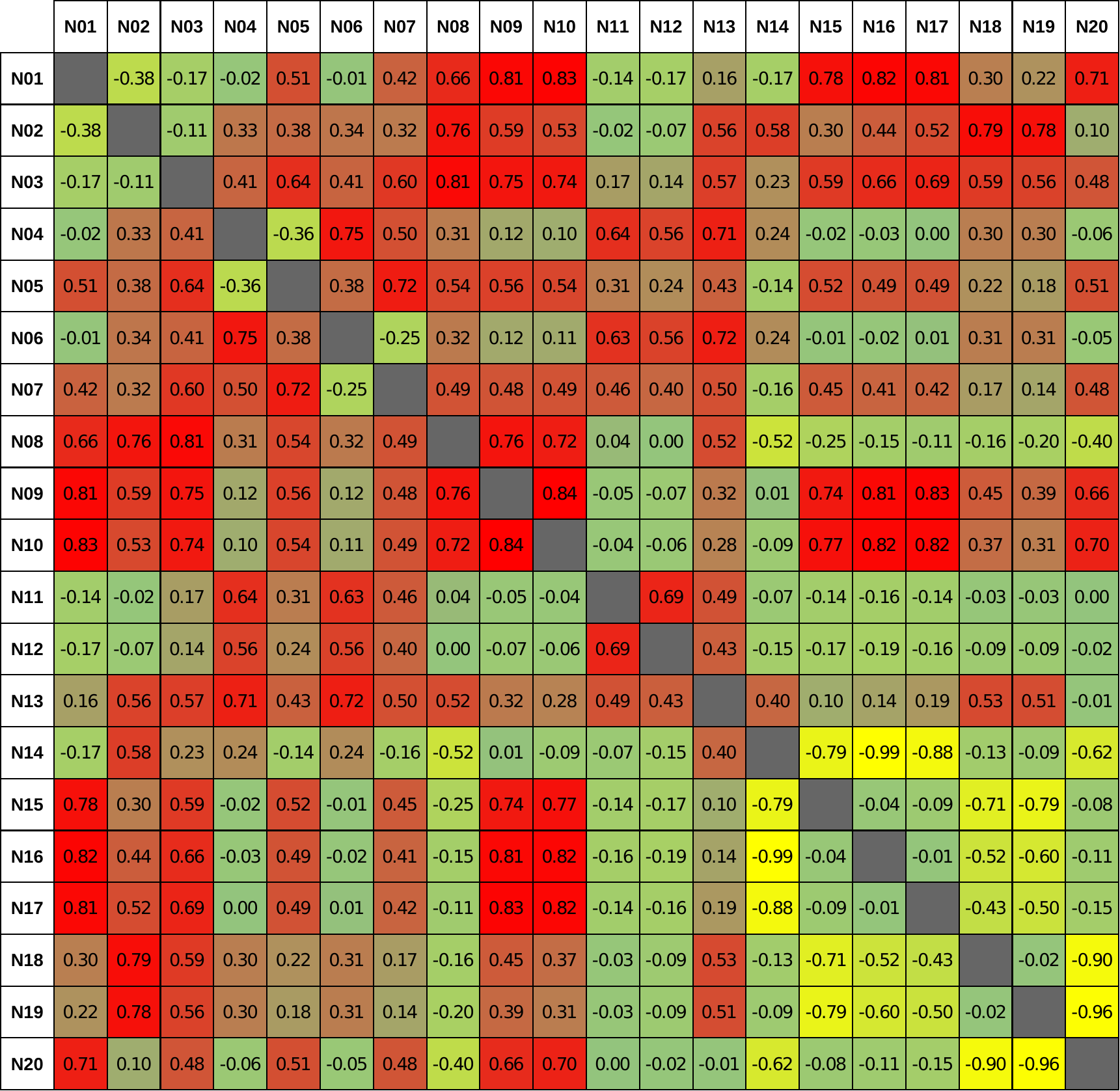}
    \caption{A pairwise comparison of twenty networks. Each cell represents the difference between two measures of visual intelligence compatibility and intrinsic networks similarity. Value \textbf{0} (\textbf{green} cells) indicates that both networks are identical and so is their visual intelligence. Value {\boldmath$-$}\textbf{1} (\textbf{yellow} cells) denotes kernels' weights are identical, yet the visual intelligence is different. Value \textbf{1} (\textbf{red} cells) signifies the opposite, two networks are distinct, yet their visual intelligence is identical.}
    \label{fig:cifar10_diff_sup}
\end{figure*}

Figure \ref{fig:cifar10_diff_sup} is the difference between these two measures.

\section{\textit{ImageNet}}
\begin{figure*}[ht]
    \centering
    \includegraphics[width=2\columnwidth]{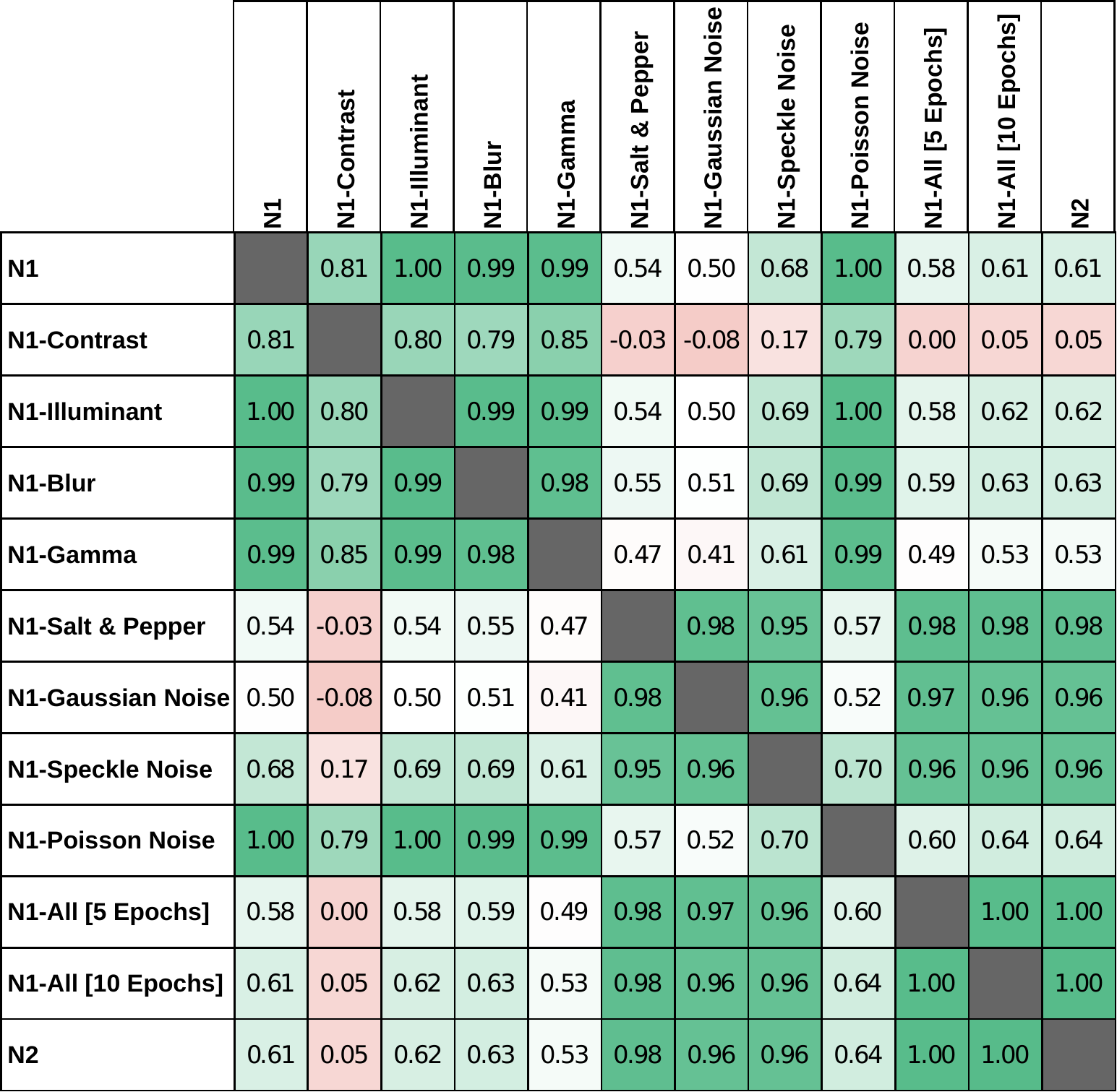}
    \caption{Pairwise comparison of visual intelligence compatibility between twelve networks. All networks are of \textit{ResNet50} architecture and have been trained on \textit{ImageNet} dataset. \textbf{Green} cells indicate high compatibility between performance of a pair of networks across all eight image distortions. \textbf{Red} cells denote the opposite, no correlation between the performance of two networks.}
    \label{fig:imagenet_vic_sup}
\end{figure*}

Figure \ref{fig:imagenet_vic_sup} illustrates visual intelligence compatibility of a subset of studied \textit{ResNet50}.

\begin{figure*}[ht]
    \centering
    \includegraphics[width=2\columnwidth]{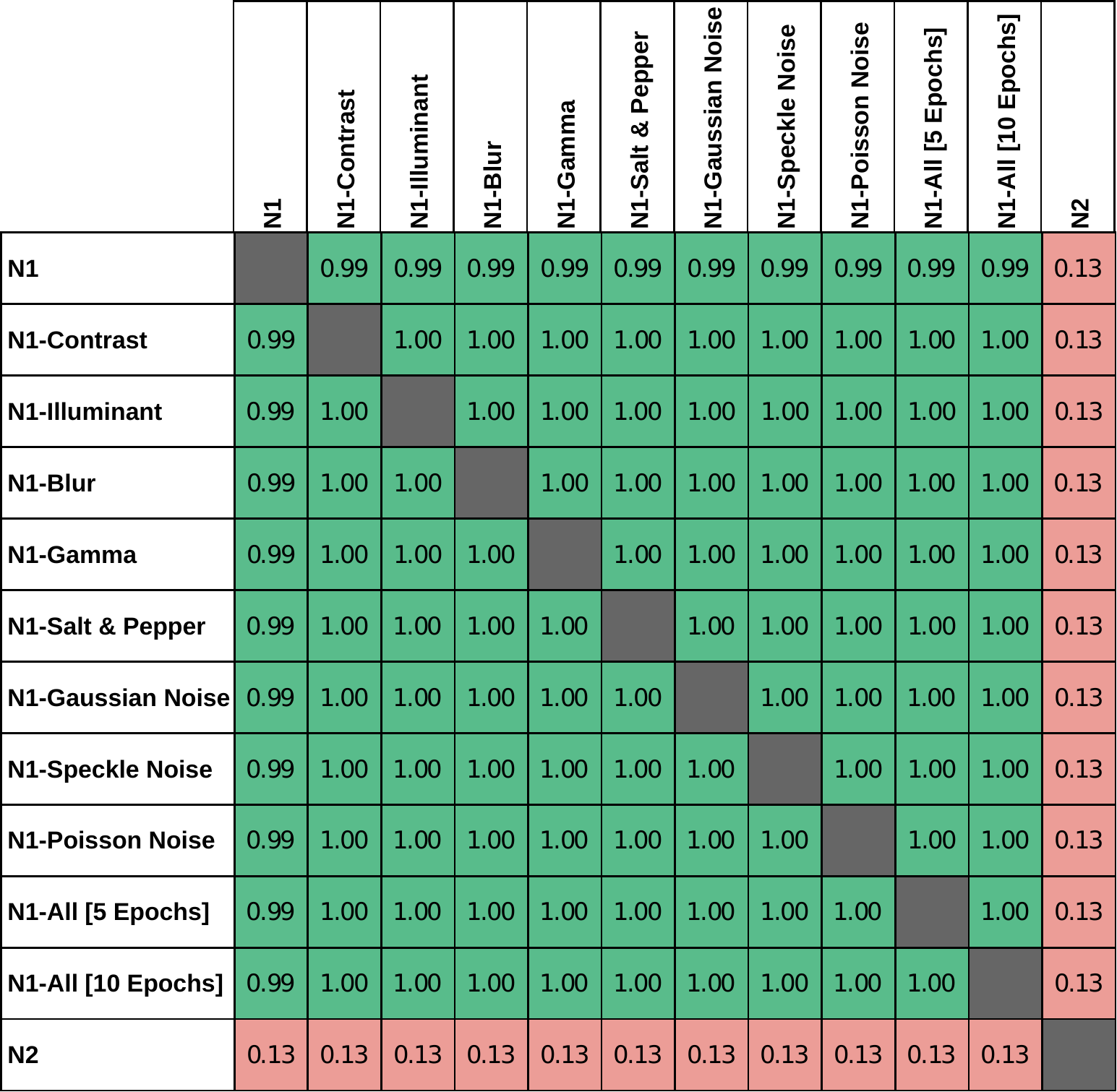}
    \caption{Pairwise comparison of intrinsic similarity between twelve networks. All networks are of \textit{ResNet50} architecture and have been trained on \textit{ImageNet} dataset. \textbf{Green} cells indicate high Pearson correlation coefficient between kernels' weights of a pair of networks. \textbf{Red} cells denote the opposite, no correlation between kernels' weights of two networks.}
    \label{fig:imagenet_int_sup}
\end{figure*}

Figure \ref{fig:imagenet_int_sup} illustrates intrinsic similarity of a subset of studied \textit{ResNet50}.

\begin{figure*}[ht]
    \centering
    \includegraphics[width=2\columnwidth]{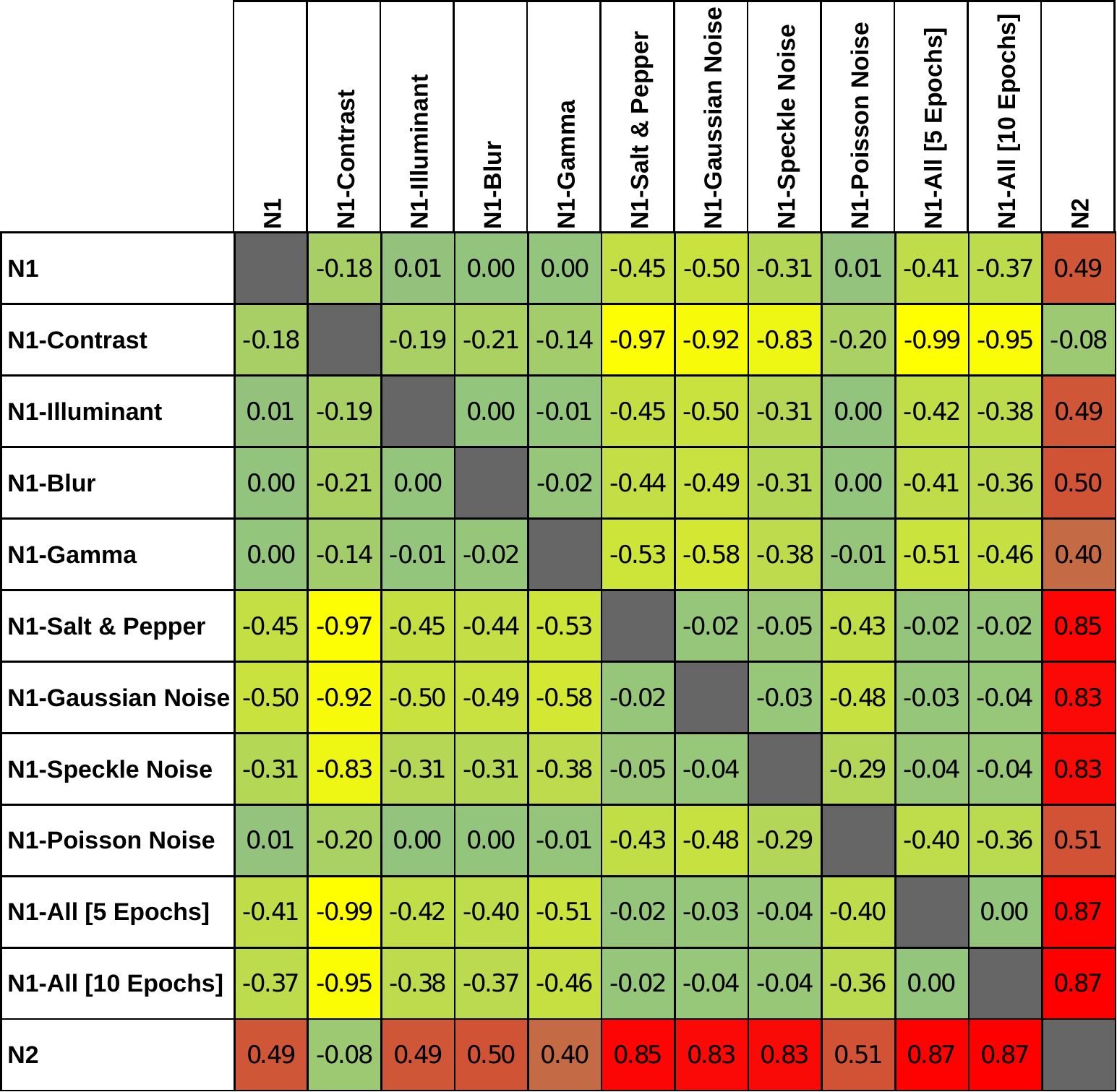}
    \caption{A pairwise comparison of twelve \textit{ResNet50}. Each cell represents the difference between two measures of visual intelligence compatibility and intrinsic networks similarity. Value \textbf{0} (\textbf{green} cells) indicates that both networks are identical and so is their visual intelligence. Value {\boldmath$-$}\textbf{1} (\textbf{yellow} cells) denotes kernels' weights are identical, yet the visual intelligence is different. Value \textbf{1} (\textbf{red} cells) signifies the opposite, two networks are distinct, yet their visual intelligence is identical.}
    \label{fig:imagenet_diff_sup}
\end{figure*}

Figure \ref{fig:imagenet_diff_sup} is the difference between these two measures.

\end{document}